\pdfoutput=1

\documentclass[11pt]{article}

\usepackage[]{acl}

\usepackage{times}
\usepackage{latexsym}

\usepackage[T1]{fontenc}

\usepackage[utf8]{inputenc}

\usepackage{microtype}

\usepackage{inconsolata}

\usepackage{xspace}
\makeatletter
\DeclareRobustCommand\onedot{\futurelet\@let@token\@onedot}
\def\@onedot{\ifx\@let@token.\else.\null\fi\xspace}

\def\eg{\emph{e.g}\onedot} 
\def\ie{\emph{i.e}\onedot}

\makeatother

\usepackage{multirow}
\usepackage{amsmath,amsthm,amssymb}
\usepackage{algorithm}
\usepackage{algpseudocode}
\usepackage{adjustbox}
\usepackage{blindtext}
\usepackage{hyperref}
\usepackage{dsfont}
\usepackage{comment}
\usepackage{subfig} 
\usepackage{booktabs,tabularx, colortbl}

\usepackage{array}
\newcolumntype{M}[1]{>{\centering\arraybackslash}m{#1}}
\newcolumntype{P}[1]{>{\centering\arraybackslash}p{#1}}

\usepackage{color}
\definecolor{P}{rgb}{0.60,0.24,0.8}
\definecolor{my_color1}{rgb}{0.94,0.87,0.8}
\definecolor{my_color2}{rgb}{0.63,0.79,0.95}
\definecolor{my_color3}{rgb}{1.0,0.79,0.95}
\definecolor{color_red}{RGB}{233,36,79}
\definecolor{color_yellow}{RGB}{200, 159, 10}
\definecolor{color_green}{RGB}{34, 139, 34}
\definecolor{color_gray}{RGB}{34, 34, 34}
\definecolor{Line}{rgb}{.5,.5,1}
\definecolor{Note}{rgb}{.75,.35,.4}
\definecolor{Row}{rgb}{.9,.25,.4}

\usepackage{tikz}
\newcommand{\tikzxmark}{%
\tikz[scale=0.23] {
    \draw[line width=0.7,line cap=round] (0,0) to [bend left=6] (1,1);
    \draw[line width=0.7,line cap=round] (0.2,0.95) to [bend right=3] (0.8,0.05);
}}
\newcommand{\tikzcmark}{%
\tikz[scale=0.23] {
    \draw[line width=0.7,line cap=round] (0.25,0) to [bend left=10] (1,1);
    \draw[line width=0.8,line cap=round] (0,0.35) to [bend right=1] (0.23,0);
}}

\newcommand{\linenumbercolor}{\color{Row}}

\usepackage{lipsum}

\usepackage[bottom]{footmisc}
\usepackage{stfloats}
\fnbelowfloat 

\title{Unsupervised hard Negative Augmentation for contrastive learning}

\author{Yuxuan Shu \and Vasileios Lampos\\
  Department of Computer Science\\
  Centre for Artificial Intelligence\\
  University College London, UK\\
  \texttt{\{yuxuan.shu.22, v.lampos\}@ucl.ac.uk}}

\begin{document}
\maketitle
\begin{abstract}
We present Unsupervised hard Negative Augmentation (UNA), a method that generates synthetic negative instances based on the term frequency-inverse document frequency (TF-IDF) retrieval model. UNA uses TF-IDF scores to ascertain the perceived importance of terms in a sentence and then produces negative samples by replacing terms with respect to that. Our experiments demonstrate that models trained with UNA improve the overall performance in semantic textual similarity tasks. Additional performance gains are obtained when combining UNA with the paraphrasing augmentation. Further results show that our method is compatible with different backbone models. Ablation studies also support the choice of having a TF-IDF-driven control on negative augmentation. 
\color{black}

\end{abstract}

\section{Introduction}
Self-supervised contrastive learning (SSCL) has emerged as a potent method for a variety of linguistic tasks such as sentiment analysis, textual entailment, and similarity retrieval~\cite{gao2021simcse,yan2021consert,giorgi2021declutr,wu2020clear,corr2020fang}. Data augmentation in SSCL is oftentimes used to generate positive instances, based on an anchor instance, with representations that are trained to be more similar to each other or to the anchor itself~\cite{he2020momentum, grill2020bootstrap}. While various data augmentations are effective in supervised NLP tasks~\cite{wei2019eda, karimi2021aeda}, they appear to lose efficacy when applied within the context of contrastive learning. For example, a dropout mask outperformed established augmentations such as cropping, word deletion, and synonym replacement~\cite{gao2021simcse}. This could indicate that the properties of the chosen text augmentations are at odds with the SSCL objective, which seeks model invariance to these changes, while it should not.

By contrast, negative data augmentation has garnered much less attention.~\citet{sinha2021negative} introduced negative data augmentation for computer vision tasks. In addition,~\citet{tang2022augcse} examined as many as 16 augmentation strategies for NLP tasks with SSCL, discovering that synthetic sentences augmented with certain methods (\eg frequently introducing grammatical errors) perform better when treated as negative samples. Interestingly, their main performance gain originated from applying augmented instances as positive samples whilst including negatively augmented instances tended to degrade the overall performance, especially in Semantic Textual Similarity (STS) tasks. 

\begin{figure}[t!]
    \begin{center}
    \includegraphics[width=0.95\linewidth]{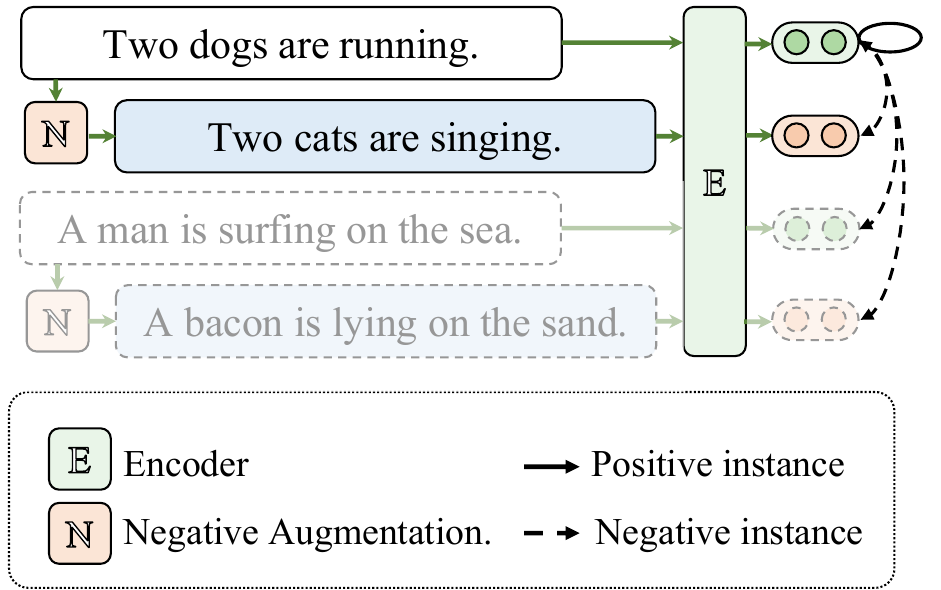} 
    \caption{Illustration of UNA. Negative augmentation, driven by TF-IDF, is applied to anchor instances to generate sentences with potential semantic differences.}
    \label{fig:intro}
    \end{center}
\end{figure}

In this paper, we argue that the potential contribution of synthetic negatives in SSCL has been undervalued. We therefore propose Unsupervised hard Negative Augmentation (UNA), an augmentation strategy for generating negative samples in SSCL.~\figureautorefname~\ref{fig:intro} depicts how UNA is applied during the model pre-training stage (details in section~\ref{sec:UNA}). UNA's core premise is the deployment of a TF-IDF-driven methodology to generate negative pairs. This improves model performance in downstream STS tasks, with additional gains when applied in conjunction with paraphrasing.

\section{Preliminaries}

\subsection{Contrastive learning} 
Contrastive learning is a training method that uses a Siamese neural network structure to learn transferable representations~\citep{das2016together,he2020momentum,chen2020a}. The key idea is to bring similar data samples closer to each other while pushing dissimilar ones apart. Given a batch of $B$ training sentences, let $x_i$, $x_j$, and $x_k$ be the anchor, positive, and negative sentences, respectively. Positive instances are defined as different views of the anchor (usually generated from different augmentations), while negative instances are other data points in the training batch. We denote their corresponding embeddings with $\mathbf{h}_i$, $\mathbf{h}_j$, and $\mathbf{h}_k$, respectively. The InfoNCE loss is used for self-supervised instance discrimination:
\begin{equation}
    \mathcal{L} =\mathbb{E}\left[-\log \frac{e^{f_{\text{sim}}\left(\mathbf{h}_{i}, \mathbf{h}_{j}\right)/\tau}}{\sum_{k=1}^{B} \mathds{1}_{[k \neq i,j]} e^{f_{\text{sim}}\left(\mathbf{h}_{i}, \mathbf{h}_{k}\right)/\tau}}\right] \, ,
    \label{eq:infonce}
\end{equation}
where $\tau$ is a temperature hyperparameter, and the {$f_{\text{sim}}$} function is a similarity metric (\eg cosine similarity). The objective of the loss function is to maximise the similarity between positive pairs while minimising it between negative pairs. This encourages the neural network to capture discriminative and potentially informative features for downstream tasks.

\subsection{Hard negative sampling}
Hard negative samples, which are negative samples that are challenging for the model to differentiate from the anchor instance, have proven to be effective in contrastive learning~\cite{kalantidis2020hard}. Instead of random sampling, there has been a focus on obtaining hard negatives using samples that are close to the anchor instance in the embedding space~\cite{robinson2021contrastive}. This may encourage some separation (disambiguation) between instances that may be close in the embedding space but have contrastive semantic interpretations. In addition, synthetic instances, obtained via transformations such as MixUp~\cite{zhang2018mixup} or Cutmix~\cite{yun2019cutmix}, have been used~\cite{sinha2021negative, shu2022revisiting}. This has showcased the utility of transformation-based generative samples that are not present in an observed dataset.

\section{Unsupervised hard Negative Augmentation (UNA)}
\label{sec:UNA}
Terms with a higher degree of substance (\ie words that are less common) are on average more important in determining the meaning of a sentence~\cite{ramos2003using}, thus swapping such words may generate semantically distant negatives for the model to train upon (see~\tableautorefname~\ref{tab:generated_samples}). Motivated by~\citet{xie2020advances}, who used TF-IDF~\cite{luhn1958the,jones1988a} to generate positive instances by swapping the words in the sentence that are less important, we now reverse this paradigm. Our proposed method (UNA) introduces a TF-IDF-driven generation of hard negative pairs, whereby words with more substance have a greater probability of being swapped, and more common words do not. UNA consists of 3 steps that we describe next (see also Appendix, Algorithm~\ref{alg:ssl_aug}). 

\paragraph{Step 1 --- Derive a TF-IDF representation.}
A TF-IDF score for a term $t$ in a document $d$ from a corpus $\mathcal{D}$ with $N$ documents is given by:
\begin{equation}
    \text{tf-idf}(t,d,\mathcal{D})\,=\,\text{tf}(t,d) \times \text{idf}(t,\mathcal{D}) \, ,
\label{eq:tfidf}
\end{equation}
where $\text{tf}(t,d) = \log(1+ n_t/n)$ and $\text{idf}(t,\mathcal{D}) = - \log(N_t/N)$. $n_t$ and $n$ respectively denote the count of term $t$ and the total count of terms in document $d$. $N_t$ denotes the number of documents from $\mathcal{D}$ that contain term $t$. The TF-IDF representation is held in a matrix $\mathbf{Z} \in \mathbb{R}^{N \times m}$, where $m$ is the number of terms in the vocabulary. 

\paragraph{Step 2 --- Determine which terms will be replaced at the sentence level.}
For a sentence in the corpus represented by the TF-IDF vector $\mathbf{z} \in \mathbf{Z}$, let $p_i$ denote the probability of replacing term $i$. $p_i$ is determined as follows: 
\begin{align}
    \label{eq:replace_term-p}
    p_i &= \min\left(\beta\left(z_i-\min\left(\mathbf{z}\right)\right)/C, 1\right) \, , \\
    \label{eq:replace_term}
    C   &= \frac{1}{n_z}\sum_{i=1}^{n_z}\left(z_i-\min\left(\mathbf{z}\right)\right) \, ,  
\end{align}
where $z_i$ is the $i$-th element of $\mathbf{z}$, $n_z$ is the number of terms that are present in the sentence, and $\beta$ is a hyperparameter that tunes the augmentation magnitude. In our experiments, we set $\beta = 0.5$ (see~\tableautorefname~\ref{tab:samplify}), and thus $p_i \in [0,1]$ with a mean of $0.5$ (See~\equationautorefname~\ref{eq:mean05}). Note that we always set $p_i = 1$ for the term with the greatest TF-IDF score in a sentence. The intuition behind this is to have at least one term replaced in every sentence, which guarantees that all augmented negative samples are different from their paired original sentences.

\paragraph{Step 3 --- Sample replacement terms based on the level of information.}
We aim to choose replacement terms that have a similar level of importance to the terms that are replacing (based on Step 2) with respect to their TF-IDF representation in our corpus. To perform this, we rely on the maximum TF-IDF score of terms in the corpus. For a term $j$, this is denoted by $s_j$, which is the maximum value of the $j$-th column of $\mathbf{Z}$. To replace term $i$ we then sample from a set of terms located in the maximum TF-IDF score vicinity of term $i$. We determine a radius of $r$ terms (below and above the reference maximum TF-IDF score of term $i$) and sample a replacement term with respect to their maximum TF-IDF scores. By encouraging replacements of terms with alternatives that have a comparable level of importance in a sentence, we generate new sentences that are more likely to retain the original structure but convey something semantically distant. Hence, these augmented sentences, guided by the TF-IDF representation $\mathbf{Z}$, could be considered as hard negative samples.

\section{Experiments and results}
\label{sec:experiments}

\paragraph{Augmentation implementation.}
UNA is applied during the self-supervised pre-training stage.\footnote{The source code is available here: \url{https://github.com/ClaudiaShu/UNA}} For $B$ sentences in a training batch, we generate $B$ negative sentences with UNA, doubling the batch size. The overall information is not doubled as UNA only generates in-distribution data. In our experiment setup, this process is carried out once every $5$ training batches. The TF-IDF representation has a vocabulary size of $451{,}691$ terms, which is obtained from the training dataset (see~\appendixautorefname~\ref{sec:dataset}). The only text preprocessing operations performed are basic tokenisation and lowercasing.

In addition, we incorporate UNA along with paraphrasing, where positive instances are generated by utilising a paraphrasing model, namely PEGASUS~\cite{zhang2020pegasus} (see~\appendixautorefname~\ref{sec:aug_detail}). Paraphrasing rephrases the anchor sentence while maintaining its original meaning. All sentences in the training corpus are paraphrased to be considered as the corresponding positive instances to the original sentences. As a result, this process doubles the amount of sentences for TF-IDF, increasing the vocabulary size to $474{,}653$ terms.

\paragraph{Baseline model.}
We first reproduce SimCSE following the self-supervised pre-training and evaluation setup in~\citet{gao2021simcse}. We note that there is a difference in performance in the reproduced result compared to the original one (while maintaining the same seed), specifically from $0.825$~(reported) to $0.816$~(reproduced) with BERT and from $0.8376$~(reported) to $0.8471$~(reproduced) with RoBERTa on the STS-B development dataset~\cite{cer2017semeval}. For a more consistent evaluation, we present all augmentation results in comparison to the reproduced version of the unsupervised SimCSE-BERT$_\texttt{base}$ and SimCSE-RoBERTa$_\texttt{base}$ model. We also run experiments with different seeds to establish a more robust difference in performance (see~\tableautorefname~\ref{tab:seed}).

\paragraph{Self-supervised pre-training.}
For the self-supervised pre-training stage, we train on the same dataset as in the paper that introduced SimCSE~\cite{gao2021simcse}. We use BERT and RoBERTa as our base models and start training with pre-trained checkpoints. To be specific, the uncased BERT model\footnote{Pre-trained BERT model, \url{https://huggingface.co/bert-base-uncased}} and the cased pre-trained RoBERTa model\footnote{Pre-trained RoBERTa model, \url{https://huggingface.co/roberta-base}}. The training batch size is set to $64$ and the temperature, $\tau$, is equal to $0.05$. Dropout is applied to all models, value set to $0.1$. The model is trained on the dataset for $1$ epoch using the AdamW optimiser~\cite{loshchilov2018decoupled}. The initial learning rate of BERT$_\texttt{base}$ models is $3$$\times$$10^{-5}$ and decays to $0$ with a step-wise linear decay scheduler (applied after each batch). The learning rate for RoBERTa$_\texttt{base}$ models is set constantly to $5$$\times$$10^{-6}$ (this improves performance compared to the one reported by~\citet{gao2021simcse}). The radius $r$ (for replacement term sampling based on the maximum TF-IDF score) is set to $4{,}000$ (\tableautorefname~\ref{tab:size} justifies this choice), which is around $1\%$ of the vocabulary size. During the self-supervised pre-training stage, we evaluate the model on the STS-B development dataset every $100$ batches. The checkpoint with the best validation result is used for downstream task evaluation.

\paragraph{Downstream task evaluation.}
The evaluation is based on 7 STS tasks, each containing pairs of sentences with corresponding labelled similarity scores. Details of the datasets can be found in~\appendixautorefname~\ref{sec:dataset}. The results are based on the unified setting provided by~\citet{gao2021simcse}. First, we freeze the pre-trained network to extract sentence embeddings from the evaluation datasets. These embeddings are obtained from the first layer of the last hidden state. Next, we compute cosine similarity scores between the embeddings of sentence pairs in the evaluation datasets. Spearman's rank correlation is used to measure the relationship between the predicted similarity scores and the ground truth.

\begin{table*}[!t]
    \small
    \centering
    \setlength{\tabcolsep}{5pt}
    \begin{tabular}{@{}ccccccccccc@{}}
    \toprule
     & \multicolumn{2}{c}{\textbf{Augmentation}} & \multicolumn{8}{c}{\textbf{SentEval \textcolor{Line}{- STS tasks }}} \\
    \midrule
    & paraphrasing & UNA & \textbf{STS12} & \textbf{STS13} & \textbf{STS14} & \textbf{STS15} & \textbf{STS16} & \textbf{STS Bench.} & \textbf{SICK Rel.} & \textbf{Avg.} \\
    \cmidrule(lr){2-3} \cmidrule(lr){4-11}
    \multirow{4}{*}[-0.5em]{\rotatebox{90}{BERT}} & \tikzxmark & \tikzxmark & .6904 & .7952 & .7303 & .8019 & .7815 & .7616 & .7116 & .7532 \\
    & \tikzxmark & \tikzcmark & .6802 & \textbf{.8226} & .7306 & .8200 & .7908 & .7760 & .7093 & .7614 \\
    \cmidrule(lr){4-11}
    & \tikzcmark & \tikzxmark & \textbf{.7003} & .7978 & .7531 & .8253 & .7908 & .7856 & .7165 & .7671 \\
    & \tikzcmark & \tikzcmark & .6939 & .8158 & \textbf{.7616} & \textbf{.8304} & \textbf{.7910} & \textbf{.8030} & \textbf{.7412} & \textbf{.7767} \\
    \midrule
    \multirow{4}{*}[-0.5em]{\rotatebox{90}{RoBERTa}}& \tikzxmark & \tikzxmark & .6907 & .8175 & .7336 & .8193 & .7974 & .7996 & .6961 & .7649 \\
    & \tikzxmark & \tikzcmark & .7075 & .8155 & .7343 & .8261 & .8053 & .8021 & .6808 & .7674 \\
    \cmidrule(lr){4-11}
    & \tikzcmark & \tikzxmark & .6962 & .8162 & .7442 & .8321 & .8000 & .8128 & \textbf{.7362} & .7768 \\
    & \tikzcmark & \tikzcmark & \textbf{.7095} & \textbf{.8188} & \textbf{.7493} & \textbf{.8334} & \textbf{.8092} & \textbf{.8189} & .7352 & \textbf{.7820} \\
    \bottomrule
    \end{tabular}
    \caption{Performance on the SemEval STS tasks (Spearman correlation) when UNA and/or paraphrasing are added with BERT and RoBERTa as the backbone model. Please note that STS Bench. and SICK Rel. abbreviate STS Benchmark and SICK Relatedness, respectively.}
    \label{tab:sts_tfidf}
\end{table*}

\paragraph{Results.} We compare adding UNA to two baselines, SimCSE with BERT$_\texttt{base}$ and RoBERTa$_\texttt{base}$. As shown in \tableautorefname~\ref{tab:sts_tfidf}, the application of UNA improves the average correlation of the unsupervised SimCSE-BERT$_\texttt{base}$ from $.7532$ to $.7614$, and from $.7649$ to $.7674$ on the unsupervised SimCSE-RoBERTa$_\texttt{base}$. Further, when both paraphrasing and UNA are deployed, correlation increases to $.7767$ with BERT backbone (an outcome which is robust across different seeds; see~\tableautorefname~\ref{tab:seed}) and further to $.7820$ with RoBERTa backbone. This is also an improvement over the sole deployment of paraphrasing ($\rho = .7671$ with BERT and $\rho = .7768$ with RoBERTa). The performance on different backbone models points to the same result pattern, demonstrating the effectiveness of UNA with a different backbone language model, and, to some extent, supporting further generalisation claims. We argue that while paraphrasing and UNA both add synthetic data points to the training data, they introduce complimentary invariances to the model. Paraphrasing contributes to recognising sentences with similar semantics, while UNA provides challenging negative signals for separating sentences, and both work in tandem. Interestingly, we reach SOTA performance on the SICK Relatedness task (unsupervised, comparison with other models in~\tableautorefname~\ref{tab:comp}), showcasing that the combination of paraphrasing with UNA can be a straightforward method to improve performance on downstream tasks. 

\begin{table}[!t]
    \small
    \centering
    \begin{tabular}{M{0.5in}M{0.5in}cc}
    \toprule
        \bf Step 2 \newline Random & \bf Step 3 \newline Random & \bf STS-B dev. & \bf STS Avg. \\
        \midrule
        \tikzcmark & \tikzcmark & .7798 & .7258 \\
        \tikzxmark & \tikzcmark & .8086 & .7384 \\
        \tikzxmark & \tikzxmark & \textbf{.8235} & \textbf{.7614} \\
    \bottomrule
    \end{tabular}
    \caption{Ablation study on negative sampling. The first column indicates whether sentence terms are randomly selected to be replaced and the second column whether the replacement term selection is random. The last row is UNA. Performance on the STS-B development set and the average correlation on the 7 STS tasks are shown.}
    \label{tab:ablation}
\end{table}

\paragraph{Ablation.} The effectiveness of UNA relies on two vital steps: selecting terms with high TF-IDF scores and replacing them with terms that have similar maximum TF-IDF scores in the corpus. Randomly swapping the terms in a sentence does not contribute to generating informative negative samples. To demonstrate this, we conducted an ablation study with BERT$_\texttt{base}$ model, validating performance on the STS-B development set as well as the average 7 STS tasks. Results are enumerated in~\tableautorefname~\ref{tab:ablation}. This highlights that guiding the term selection and replacement process with TF-IDF provides significantly better outcomes than doing either or both steps randomly.

\section{Conclusion}
In this paper, we propose a novel and efficient negative augmentation strategy, UNA, guided by the TF-IDF retrieval model. Results show that UNA is compatible with different backbone models and could bring great performance improvements to downstream STS tasks, especially when combined with paraphrasing.

\section{Limitations}
Our experiments focused on English language datasets. Therefore, the findings may not provide constructive directions for other languages with different characteristics compared to English. In addition, the application of TF-IDF on single sentences (as opposed to lengthier text constructs) might provide quite dataset-specific results which may or may not work in favour of the proposed augmentation method (UNA). This is something that follow-up work should investigate.

\section*{Acknowledgements} VL would like to acknowledge all levels of support from the EPSRC grant EP/X031276/1.

\bibliography{anthology}

\begin{thebibliography}{42}
\expandafter\ifx\csname natexlab\endcsname\relax\def\natexlab#1{#1}\fi

\bibitem[{Agirre et~al.(2015)Agirre, Banea, Cardie, Cer, Diab, Gonzalez-Agirre, Guo, Lopez-Gazpio, Maritxalar, Mihalcea, Rigau, Uria, and Wiebe}]{agirre2015semeval}
Eneko Agirre, Carmen Banea, Claire Cardie, Daniel Cer, Mona Diab, Aitor Gonzalez-Agirre, Weiwei Guo, I{\~n}igo Lopez-Gazpio, Montse Maritxalar, Rada Mihalcea, German Rigau, Larraitz Uria, and Janyce Wiebe. 2015.
\newblock \href {https://doi.org/10.18653/v1/S15-2045} {{SemEval-2015 Task 2: Semantic Textual Similarity, English, Spanish and Pilot on Interpretability}}.
\newblock In \emph{Proceedings of the 9th International Workshop on Semantic Evaluation ({S}em{E}val 2015)}, pages 252--263.

\bibitem[{Agirre et~al.(2014)Agirre, Banea, Cardie, Cer, Diab, Gonzalez-Agirre, Guo, Mihalcea, Rigau, and Wiebe}]{agirre2014semeval}
Eneko Agirre, Carmen Banea, Claire Cardie, Daniel Cer, Mona Diab, Aitor Gonzalez-Agirre, Weiwei Guo, Rada Mihalcea, German Rigau, and Janyce Wiebe. 2014.
\newblock \href {https://doi.org/10.3115/v1/S14-2010} {{SemEval-2014 Task 10: Multilingual Semantic Textual Similarity}}.
\newblock In \emph{Proceedings of the 8th International Workshop on Semantic Evaluation ({S}em{E}val 2014)}, pages 81--91.

\bibitem[{Agirre et~al.(2016)Agirre, Banea, Cer, Diab, Gonzalez-Agirre, Mihalcea, Rigau, and Wiebe}]{agirre2016semeval}
Eneko Agirre, Carmen Banea, Daniel Cer, Mona Diab, Aitor Gonzalez-Agirre, Rada Mihalcea, German Rigau, and Janyce Wiebe. 2016.
\newblock \href {https://doi.org/10.18653/v1/S16-1081} {{SemEval-2016 Task 1: Semantic Textual Similarity, Monolingual and Cross-Lingual Evaluation}}.
\newblock In \emph{Proceedings of the 10th International Workshop on Semantic Evaluation ({S}em{E}val-2016)}, pages 497--511.

\bibitem[{Agirre et~al.(2012)Agirre, Cer, Diab, and Gonzalez-Agirre}]{agirre2012semeval}
Eneko Agirre, Daniel Cer, Mona Diab, and Aitor Gonzalez-Agirre. 2012.
\newblock \href {https://aclanthology.org/S12-1051} {{SemEval-2012 Task 6: A Pilot on Semantic Textual Similarity}}.
\newblock In \emph{*{SEM} 2012: The First Joint Conference on Lexical and Computational Semantics {--} Volume 1: Proceedings of the main conference and the shared task, and Volume 2: Proceedings of the Sixth International Workshop on Semantic Evaluation ({S}em{E}val 2012)}, pages 385--393.

\bibitem[{Agirre et~al.(2013)Agirre, Cer, Diab, Gonzalez-Agirre, and Guo}]{agirre2013sem}
Eneko Agirre, Daniel Cer, Mona Diab, Aitor Gonzalez-Agirre, and Weiwei Guo. 2013.
\newblock \href {https://aclanthology.org/S13-1004} {{*SEM 2013 shared task: Semantic Textual Similarity}}.
\newblock In \emph{Second Joint Conference on Lexical and Computational Semantics (*{SEM}), Volume 1: Proceedings of the Main Conference and the Shared Task: Semantic Textual Similarity}, pages 32--43.

\bibitem[{Cer et~al.(2017)Cer, Diab, Agirre, Lopez-Gazpio, and Specia}]{cer2017semeval}
Daniel Cer, Mona Diab, Eneko Agirre, I{\~n}igo Lopez-Gazpio, and Lucia Specia. 2017.
\newblock \href {https://doi.org/10.18653/v1/S17-2001} {{SemEval-2017 Task 1: Semantic Textual Similarity Multilingual and Crosslingual Focused Evaluation}}.
\newblock In \emph{Proceedings of the 11th International Workshop on Semantic Evaluation ({S}em{E}val-2017)}, pages 1--14.

\bibitem[{Chen et~al.(2020)Chen, Kornblith, Norouzi, and Hinton}]{chen2020a}
Ting Chen, Simon Kornblith, Mohammad Norouzi, and Geoffrey Hinton. 2020.
\newblock \href {https://proceedings.mlr.press/v119/chen20j.html} {{A Simple Framework for Contrastive Learning of Visual Representations}}.
\newblock In \emph{Proceedings of the 37th International Conference on Machine Learning}, volume 119, pages 1597--1607.

\bibitem[{Cheung and Yeung(2021)}]{cheung2021modals}
Tsz-Him Cheung and Dit-Yan Yeung. 2021.
\newblock \href {https://openreview.net/forum?id=XjYgR6gbCEc} {{MODALS: Modality-agnostic Automated Data Augmentation in the Latent Space}}.
\newblock In \emph{International Conference on Learning Representations}.

\bibitem[{Chuang et~al.(2022)Chuang, Dangovski, Luo, Zhang, Chang, Soljacic, Li, Yih, Kim, and Glass}]{chuang2022diffcse}
Yung-Sung Chuang, Rumen Dangovski, Hongyin Luo, Yang Zhang, Shiyu Chang, Marin Soljacic, Shang-Wen Li, Scott Yih, Yoon Kim, and James Glass. 2022.
\newblock \href {https://doi.org/10.18653/v1/2022.naacl-main.311} {{D}iff{CSE}: Difference-based contrastive learning for sentence embeddings}.
\newblock In \emph{Proceedings of the 2022 Conference of the North American Chapter of the Association for Computational Linguistics: Human Language Technologies}, pages 4207--4218.

\bibitem[{Das et~al.(2016)Das, Yenala, Chinnakotla, and Shrivastava}]{das2016together}
Arpita Das, Harish Yenala, Manoj Chinnakotla, and Manish Shrivastava. 2016.
\newblock \href {https://doi.org/10.18653/v1/P16-1036} {{Together we stand: Siamese Networks for Similar Question Retrieval}}.
\newblock In \emph{Proceedings of the 54th Annual Meeting of the Association for Computational Linguistics (Volume 1: Long Papers)}, pages 378--387.

\bibitem[{Fang and Xie(2020)}]{corr2020fang}
Hongchao Fang and Pengtao Xie. 2020.
\newblock \href {https://arxiv.org/abs/2005.12766} {{CERT: Contrastive Self-supervised Learning for Language Understanding}}.
\newblock \emph{arXiv preprint 2005.12766}.

\bibitem[{Gao et~al.(2021)Gao, Yao, and Chen}]{gao2021simcse}
Tianyu Gao, Xingcheng Yao, and Danqi Chen. 2021.
\newblock \href {https://doi.org/10.18653/v1/2021.emnlp-main.552} {{SimCSE: Simple Contrastive Learning of Sentence Embeddings}}.
\newblock In \emph{Proceedings of the 2021 Conference on Empirical Methods in Natural Language Processing}, pages 6894--6910.

\bibitem[{Giorgi et~al.(2021)Giorgi, Nitski, Wang, and Bader}]{giorgi2021declutr}
John Giorgi, Osvald Nitski, Bo~Wang, and Gary Bader. 2021.
\newblock \href {https://doi.org/10.18653/v1/2021.acl-long.72} {{DeCLUTR: Deep Contrastive Learning for Unsupervised Textual Representations}}.
\newblock In \emph{Proceedings of the 59th Annual Meeting of the Association for Computational Linguistics and the 11th International Joint Conference on Natural Language Processing (Volume 1: Long Papers)}, pages 879--895.

\bibitem[{Grill et~al.(2020)Grill, Strub, Altch\'{e}, Tallec, Richemond, Buchatskaya, Doersch, Avila~Pires, Guo, Gheshlaghi~Azar, Piot, kavukcuoglu, Munos, and Valko}]{grill2020bootstrap}
Jean-Bastien Grill, Florian Strub, Florent Altch\'{e}, Corentin Tallec, Pierre Richemond, Elena Buchatskaya, Carl Doersch, Bernardo Avila~Pires, Zhaohan Guo, Mohammad Gheshlaghi~Azar, Bilal Piot, koray kavukcuoglu, Remi Munos, and Michal Valko. 2020.
\newblock \href {https://proceedings.neurips.cc/paper_files/paper/2020/file/f3ada80d5c4ee70142b17b8192b2958e-Paper.pdf} {Bootstrap your own latent - a new approach to self-supervised learning}.
\newblock In \emph{Advances in Neural Information Processing Systems}, volume~33, pages 21271--21284.

\bibitem[{He et~al.(2020)He, Fan, Wu, Xie, and Girshick}]{he2020momentum}
Kaiming He, Haoqi Fan, Yuxin Wu, Saining Xie, and Ross Girshick. 2020.
\newblock \href {https://openaccess.thecvf.com/content_CVPR_2020/papers/He_Momentum_Contrast_for_Unsupervised_Visual_Representation_Learning_CVPR_2020_paper.pdf} {{Momentum Contrast for Unsupervised Visual Representation Learning}}.
\newblock In \emph{Proceedings of the IEEE/CVF Conference on Computer Vision and Pattern Recognition (CVPR)}, pages 9729--9738.

\bibitem[{Kalantidis et~al.(2020)Kalantidis, Sariyildiz, Pion, Weinzaepfel, and Larlus}]{kalantidis2020hard}
Yannis Kalantidis, Mert~Bulent Sariyildiz, Noe Pion, Philippe Weinzaepfel, and Diane Larlus. 2020.
\newblock \href {https://proceedings.neurips.cc/paper_files/paper/2020/file/f7cade80b7cc92b991cf4d2806d6bd78-Paper.pdf} {{Hard Negative Mixing for Contrastive Learning}}.
\newblock In \emph{Advances in Neural Information Processing Systems}, volume~33, pages 21798--21809.

\bibitem[{Karimi et~al.(2021)Karimi, Rossi, and Prati}]{karimi2021aeda}
Akbar Karimi, Leonardo Rossi, and Andrea Prati. 2021.
\newblock \href {https://doi.org/10.18653/v1/2021.findings-emnlp.234} {{AEDA}: An easier data augmentation technique for text classification}.
\newblock In \emph{Findings of the Association for Computational Linguistics: EMNLP 2021}, pages 2748--2754.

\bibitem[{Klein and Nabi(2022)}]{klein2022scd}
Tassilo Klein and Moin Nabi. 2022.
\newblock \href {https://doi.org/10.18653/v1/2022.acl-short.44} {{SCD}: Self-contrastive decorrelation of sentence embeddings}.
\newblock In \emph{Proceedings of the 60th Annual Meeting of the Association for Computational Linguistics (Volume 2: Short Papers)}, pages 394--400.

\bibitem[{Liu et~al.(2021)Liu, Vuli{\'c}, Korhonen, and Collier}]{liu2021fast}
Fangyu Liu, Ivan Vuli{\'c}, Anna Korhonen, and Nigel Collier. 2021.
\newblock \href {https://doi.org/10.18653/v1/2021.emnlp-main.109} {Fast, effective, and self-supervised: Transforming masked language models into universal lexical and sentence encoders}.
\newblock In \emph{Proceedings of the 2021 Conference on Empirical Methods in Natural Language Processing}, pages 1442--1459.

\bibitem[{Loshchilov and Hutter(2019)}]{loshchilov2018decoupled}
Ilya Loshchilov and Frank Hutter. 2019.
\newblock \href {https://openreview.net/forum?id=Bkg6RiCqY7} {{Decoupled Weight Decay Regularization}}.
\newblock In \emph{International Conference on Learning Representations}.

\bibitem[{Luhn(1958)}]{luhn1958the}
Hans~Peter Luhn. 1958.
\newblock \href {https://doi.org/10.1147/rd.22.0159} {{The Automatic Creation of Literature Abstracts}}.
\newblock \emph{IBM Journal of Research and Development}, 2(2):159--165.

\bibitem[{Marelli et~al.(2014)Marelli, Menini, Baroni, Bentivogli, Bernardi, and Zamparelli}]{marelli2014sick}
Marco Marelli, Stefano Menini, Marco Baroni, Luisa Bentivogli, Raffaella Bernardi, and Roberto Zamparelli. 2014.
\newblock \href {http://www.lrec-conf.org/proceedings/lrec2014/pdf/363_Paper.pdf} {{A SICK cure for the evaluation of compositional distributional semantic models}}.
\newblock In \emph{Proceedings of the Ninth International Conference on Language Resources and Evaluation ({LREC}'14)}, pages 216--223.

\bibitem[{Miller(1995)}]{miller1999wordnet}
George~A. Miller. 1995.
\newblock \href {https://doi.org/10.1145/219717.219748} {{WordNet: A Lexical Database for English}}.
\newblock \emph{Communications of the ACM}, 38(11):39--41.

\bibitem[{Paszke et~al.(2019)Paszke, Gross, Massa, Lerer, Bradbury, Chanan, Killeen, Lin, Gimelshein, Antiga, Desmaison, Kopf, Yang, DeVito, Raison, Tejani, Chilamkurthy, Steiner, Fang, Bai, and Chintala}]{paszke2019pytorch}
Adam Paszke, Sam Gross, Francisco Massa, Adam Lerer, James Bradbury, Gregory Chanan, Trevor Killeen, Zeming Lin, Natalia Gimelshein, Luca Antiga, Alban Desmaison, Andreas Kopf, Edward Yang, Zachary DeVito, Martin Raison, Alykhan Tejani, Sasank Chilamkurthy, Benoit Steiner, Lu~Fang, Junjie Bai, and Soumith Chintala. 2019.
\newblock \href {http://papers.neurips.cc/paper/9015-pytorch-an-imperative-style-high-performance-deep-learning-library.pdf} {{PyTorch: An Imperative Style, High-Performance Deep Learning Library}}.
\newblock In \emph{Advances in Neural Information Processing Systems}, volume~32, pages 8024--8035.

\bibitem[{Raffel et~al.(2020)Raffel, Shazeer, Roberts, Lee, Narang, Matena, Zhou, Li, and Liu}]{colin2020exploring}
Colin Raffel, Noam Shazeer, Adam Roberts, Katherine Lee, Sharan Narang, Michael Matena, Yanqi Zhou, Wei Li, and Peter~J. Liu. 2020.
\newblock \href {http://jmlr.org/papers/v21/20-074.html} {{Exploring the Limits of Transfer Learning with a Unified Text-to-Text Transformer}}.
\newblock \emph{Journal of Machine Learning Research}, 21(140):1--67.

\bibitem[{Ramos et~al.(2003)}]{ramos2003using}
Juan Ramos et~al. 2003.
\newblock \href {https://citeseerx.ist.psu.edu/document?repid=rep1&type=pdf&doi=b3bf6373ff41a115197cb5b30e57830c16130c2c} {Using tf-idf to determine word relevance in document queries}.
\newblock In \emph{Proceedings of the first instructional conference on machine learning}, volume 242, pages 29--48.

\bibitem[{Robinson et~al.(2021)Robinson, Chuang, Sra, and Jegelka}]{robinson2021contrastive}
Joshua~David Robinson, Ching-Yao Chuang, Suvrit Sra, and Stefanie Jegelka. 2021.
\newblock \href {https://openreview.net/forum?id=CR1XOQ0UTh-} {{Contrastive Learning with Hard Negative Samples}}.
\newblock In \emph{International Conference on Learning Representations}.

\bibitem[{Sennrich et~al.(2016)Sennrich, Haddow, and Birch}]{sennrich2016improving}
Rico Sennrich, Barry Haddow, and Alexandra Birch. 2016.
\newblock \href {https://doi.org/10.18653/v1/P16-1009} {{Improving Neural Machine Translation Models with Monolingual Data}}.
\newblock In \emph{Proceedings of the 54th Annual Meeting of the Association for Computational Linguistics (Volume 1: Long Papers)}, pages 86--96.

\bibitem[{Shu et~al.(2022)Shu, Gu, Yang, and Lo}]{shu2022revisiting}
Yuxuan Shu, Xiao Gu, Guang-Zhong Yang, and Benny P~L Lo. 2022.
\newblock \href {https://bmvc2022.mpi-inf.mpg.de/0406.pdf} {{Revisiting Self-Supervised Contrastive Learning for Facial Expression Recognition}}.
\newblock In \emph{33rd British Machine Vision Conference 2022, BMVC 2022}.

\bibitem[{Sinha et~al.(2021)Sinha, Ayush, Song, Uzkent, Jin, and Ermon}]{sinha2021negative}
Abhishek Sinha, Kumar Ayush, Jiaming Song, Burak Uzkent, Hongxia Jin, and Stefano Ermon. 2021.
\newblock \href {https://openreview.net/forum?id=Ovp8dvB8IBH} {{Negative Data Augmentation}}.
\newblock In \emph{International Conference on Learning Representations}.

\bibitem[{Sparck~Jones(1972)}]{jones1988a}
Karen Sparck~Jones. 1972.
\newblock \href {https://www.emerald.com/insight/content/doi/10.1108/eb026526/full/html} {A statistical interpretation of term specificity and its application in retrieval}.
\newblock \emph{Journal of documentation}, 28(1):11--21.

\bibitem[{Tang et~al.(2022)Tang, Kocyigit, and Wijaya}]{tang2022augcse}
Zilu Tang, Muhammed~Yusuf Kocyigit, and Derry~Tanti Wijaya. 2022.
\newblock \href {https://aclanthology.org/2022.aacl-main.30} {{A}ug{CSE}: Contrastive sentence embedding with diverse augmentations}.
\newblock In \emph{Proceedings of the 2nd Conference of the Asia-Pacific Chapter of the Association for Computational Linguistics and the 12th International Joint Conference on Natural Language Processing (Volume 1: Long Papers)}, pages 375--398.

\bibitem[{Tiedemann and Thottingal(2020)}]{tiedemann2020opus}
J{\"o}rg Tiedemann and Santhosh Thottingal. 2020.
\newblock \href {https://aclanthology.org/2020.eamt-1.61} {{OPUS-MT -- Building open translation services for the World}}.
\newblock In \emph{Proceedings of the 22nd Annual Conference of the European Association for Machine Translation}, pages 479--480.

\bibitem[{Wei and Zou(2019)}]{wei2019eda}
Jason Wei and Kai Zou. 2019.
\newblock \href {https://doi.org/10.18653/v1/D19-1670} {{EDA: Easy Data Augmentation Techniques for Boosting Performance on Text Classification Tasks}}.
\newblock In \emph{Proceedings of the 2019 Conference on Empirical Methods in Natural Language Processing and the 9th International Joint Conference on Natural Language Processing (EMNLP-IJCNLP)}, pages 6382--6388.

\bibitem[{Wu et~al.(2020)Wu, Wang, Gu, Khabsa, Sun, and Ma}]{wu2020clear}
Zhuofeng Wu, Sinong Wang, Jiatao Gu, Madian Khabsa, Fei Sun, and Hao Ma. 2020.
\newblock \href {https://doi.org/10.48550/arXiv.2012.15466} {{CLEAR: Contrastive Learning for Sentence Representation}}.
\newblock \emph{arXiv preprint arXiv:2012.15466}.

\bibitem[{Xie et~al.(2020)Xie, Dai, Hovy, Luong, and Le}]{xie2020advances}
Qizhe Xie, Zihang Dai, Eduard Hovy, Thang Luong, and Quoc Le. 2020.
\newblock \href {https://proceedings.neurips.cc/paper/2020/file/44feb0096faa8326192570788b38c1d1-Paper.pdf} {{Unsupervised Data Augmentation for Consistency Training}}.
\newblock In \emph{Advances in Neural Information Processing Systems}, volume~33, pages 6256--6268.

\bibitem[{Yan et~al.(2021)Yan, Li, Wang, Zhang, Wu, and Xu}]{yan2021consert}
Yuanmeng Yan, Rumei Li, Sirui Wang, Fuzheng Zhang, Wei Wu, and Weiran Xu. 2021.
\newblock \href {https://doi.org/10.18653/v1/2021.acl-long.393} {{ConSERT: A Contrastive Framework for Self-Supervised Sentence Representation Transfer}}.
\newblock In \emph{Proceedings of the 59th Annual Meeting of the Association for Computational Linguistics and the 11th International Joint Conference on Natural Language Processing (Volume 1: Long Papers)}, pages 5065--5075.

\bibitem[{Ye et~al.(2021)Ye, Kim, and Oh}]{ye2021efficient}
Seonghyeon Ye, Jiseon Kim, and Alice Oh. 2021.
\newblock \href {https://doi.org/10.18653/v1/2021.emnlp-main.138} {{Efficient Contrastive Learning via Novel Data Augmentation and Curriculum Learning}}.
\newblock In \emph{Proceedings of the 2021 Conference on Empirical Methods in Natural Language Processing}, pages 1832--1838.

\bibitem[{Yun et~al.(2019)Yun, Han, Oh, Chun, Choe, and Yoo}]{yun2019cutmix}
Sangdoo Yun, Dongyoon Han, Seong~Joon Oh, Sanghyuk Chun, Junsuk Choe, and Youngjoon Yoo. 2019.
\newblock \href {https://openaccess.thecvf.com/content_ICCV_2019/papers/Yun_CutMix_Regularization_Strategy_to_Train_Strong_Classifiers_With_Localizable_Features_ICCV_2019_paper.pdf} {{CutMix: Regularization Strategy to Train Strong Classifiers With Localizable Features}}.
\newblock In \emph{Proceedings of the IEEE/CVF International Conference on Computer Vision (ICCV)}.

\bibitem[{Zhang et~al.(2018)Zhang, Cisse, Dauphin, and Lopez-Paz}]{zhang2018mixup}
Hongyi Zhang, Moustapha Cisse, Yann~N. Dauphin, and David Lopez-Paz. 2018.
\newblock \href {https://openreview.net/forum?id=r1Ddp1-Rb} {{mixup: Beyond Empirical Risk Minimization}}.
\newblock In \emph{International Conference on Learning Representations}.

\bibitem[{Zhang et~al.(2020)Zhang, Zhao, Saleh, and Liu}]{zhang2020pegasus}
Jingqing Zhang, Yao Zhao, Mohammad Saleh, and Peter Liu. 2020.
\newblock \href {https://proceedings.mlr.press/v119/zhang20ae.html} {{PEGASUS: Pre-training with Extracted Gap-sentences for Abstractive Summarization}}.
\newblock In \emph{Proceedings of the 37th International Conference on Machine Learning}, volume 119, pages 11328--11339.

\bibitem[{Zhou et~al.(2022)Zhou, Zhang, Zhao, and Wen}]{zhou2022debiased}
Kun Zhou, Beichen Zhang, Xin Zhao, and Ji-Rong Wen. 2022.
\newblock \href {https://doi.org/10.18653/v1/2022.acl-long.423} {Debiased contrastive learning of unsupervised sentence representations}.
\newblock In \emph{Proceedings of the 60th Annual Meeting of the Association for Computational Linguistics (Volume 1: Long Papers)}, pages 6120--6130.

\end{thebibliography}
\bibliographystyle{acl_natbib}

\clearpage

\appendix

\section*{Appendix}
\label{sec:appendix}

\setcounter{table}{0}
\renewcommand{\thetable}{\thesection \arabic{table}}
\setcounter{figure}{0}
\renewcommand{\thefigure}{\thesection \arabic{figure}}

\section{Dataset details}
\label{sec:dataset}

\paragraph{Self-supervised pre-training dataset.} All models are pre-trained on the English Wikipedia corpus ($1$ million sentences) provided by~\citet{gao2021simcse}. We note that while we use this training dataset for a consistent comparison with the previous work, it is not entirely suitable for obtaining a TF-IDF representation due to its very short document length, and quite restrictive topic coverage.

\paragraph{STS tasks.} The evaluation is based on 7 STS tasks, namely STS 2012-2016~\cite{agirre2012semeval,agirre2013sem,agirre2014semeval,agirre2015semeval,agirre2016semeval}, STS Benchmark~\cite{cer2017semeval}, and SICK Relatedness~\cite{marelli2014sick}, each containing pairs of sentences with their corresponding labelled similarity score ranging from $0$ to $5$. The size of each evaluation (test) set is enumerated in \tableautorefname~\ref{tab:STS-size}. 

\begin{table}[b]
  \small
  \centering
  \captionsetup[subfloat]{labelformat=empty} 
  \subfloat[]{
    \begin{tabular}{cc}
        \toprule
        \bf Task & \bf Test samples\\
        \midrule
        STS12 & 3{,}108 \\
        STS13 & 1{,}500 \\
        STS14 & 3{,}750 \\
        STS15 & 3{,}000 \\
        \bottomrule
    \end{tabular}
  }
  \subfloat[]{
    \begin{tabular}{cc}
        \toprule
        \bf Task & \bf Test samples \\
        \midrule
        STS16 & 1{,}186 \\
        STS Bench. & 1{,}379 \\
        SICK Rel. & 4{,}927 \\
        \bottomrule
        \\
    \end{tabular}
  }
  \caption{Size of STS test datasets.}
  \label{tab:STS-size}
\end{table}

\section{Methodology details}
\label{sec:method_detail}
We further demonstrate how $\beta$ controls the augmentation magnitude in UNA. According to~\equationautorefname~\ref{eq:replace_term-p} and~\ref{eq:replace_term}, if we set $a_i = z_i-\min\left(\mathbf{z}\right)$, then $p_i = \min\left(\beta a_i/C, 1\right)$. We temporarily ignore the maximum boundary of $1$ to simplify the analysis. The mean of $p$, denoted as $\bar p$, can be depicted as:
\begin{align}
    \bar{p} &= \frac{1}{n_z}\sum_{i=1}^{n_z}\left(\frac{\beta a_i}{\frac{1}{n_z}\sum_{i=1}^{n_z} a_i} \right) \, .
    \label{eq:mean05}
\end{align}
Given that $\frac{1}{n_z}\sum_{i=1}^{n_z} a_i$ is a constant value for every sentence, we have $\bar{p} = \beta$. When factoring in the upper boundary, this relationship is modified to $\bar{p} \leq \beta$, which suggests that the mean probability of replacing term $i$ in the sentence closely aligns with the augmentation magnitude $\beta$.

Examples of sentences generated with UNA are tabulated in~\tableautorefname~\ref{tab:generated_samples} using the vocabulary of the training dataset described in~\sectionautorefname~\ref{sec:dataset}.

\begin{table}[t]
    \small
    \centering
    \setlength{\tabcolsep}{3pt}
    \begin{tabular}{@{}ll@{}}
    \toprule
        \textbf{Sentence} \\
    \midrule
         \textcolor{color_green}{Another factor was caffeine.} \\
         another \textbf{instrumental} was \textbf{x-45c}.  \\
         \midrule
         \textcolor{color_green}{Greetings from the real universe.}  \\
         \textbf{acylations} from the real \textbf{government-owned}.  \\
         \midrule
         \textcolor{color_green}{We should play with legos at camp.}  \\
         we \textbf{kulish} play with \textbf{minimum} at \textbf{yakiudon}.  \\
    \bottomrule
    \end{tabular}
    \caption{UNA-generated sentences. The original sentences are in \textcolor{color_green}{green} while the sentences augmented with UNA are in black. Words that have been replaced by another word are bolded.}
    \label{tab:generated_samples}
\end{table}

\section{Augmentation details}
\label{sec:aug_detail}

\subsection{Paraphrasing} 
In our experiments, paraphrasing is applied as an augmentation strategy to create positive samples of the anchor instances. To paraphrase each sentence in the training dataset, we use a paraphrasing model from the Huggingface hub,\footnote{Fine-tuned PEGASUS model for paraphrasing, \url{https://huggingface.co/tuner007/pegasus_paraphrase}} which is fine-tuned based on the PEGASUS model~\citep{zhang2020pegasus}. During pre-training, we consider the original sentences and their corresponding paraphrased ones as positive pairs. Regarding the choice of paraphrasing model, we also assessed T5 by~\citet{colin2020exploring},\footnote{Fine-tuned T5 model for paraphrasing, \url{https://huggingface.co/Vamsi/T5_Paraphrase_Paws}} but it only reached a correlation of $.7985$ on the STS-B development dataset (compared to $.8234$ for PEGASUS).

\subsection{Other augmentations}
\begin{table*}[t]
    \small
    \centering
    \setlength{\tabcolsep}{3pt}
    \begin{tabular}{@{}lcccccccc@{}}
    \toprule
    \multirow{2}{*}{\textbf{Augmentations}} & \multicolumn{8}{c}{\textbf{SentEval \textcolor{Line}{- STS tasks }}} \\
    \cmidrule(lr){2-9} 
    &  \textbf{STS12} & \textbf{STS13} & \textbf{STS14} & \textbf{STS15} & \textbf{STS16} & \textbf{STS Bench.} & \textbf{SICK Rel.} & \textbf{Avg.} \\
    \toprule
    \textbf{SimCSE}-BERT$_\texttt{base}$$^*$ \begin{scriptsize}(reproduced)\end{scriptsize} & .6904 & .7952 & .7303 & .8019 & .7815 & .7616 & .7116 & .7532 \\
    \midrule
    \textbf{EDA}$^\spadesuit$ synonym replacement & .6118 & .7106 & .6590 & .7179 & .7424 & .6623 & .6218 & .6751 \\
    \textbf{EDA}$^\spadesuit$ random insert & .6594 & .7402 & .7039 & .7698 & .7543 & .7287 & \textbf{.7200} & .7252 \\
    \textbf{EDA}$^\spadesuit$ random swap & .6850 & .7808 & .7212 & .7830 & .7767 & .7643 & .6984 & .7442 \\
    \textbf{EDA}$^\spadesuit$ random delete & .6719 & .7842 & .7147 & .7868 & .7872 & .7641 & .6916 & .7429 \\
    \textbf{back-translation}$_\texttt{ Helsinki}$$^\heartsuit$ & .6876 & .7510 & .7072 & .7933 & .7802 & .7618 & .7034 & .7406 \\
    \textbf{paraphrasing}$_\texttt{ PEGASUS}$$^\clubsuit$ & .7003 & .7978 & \textbf{.7531} & \textbf{.8253} & \textbf{.7908} & \textbf{.7856} & .7165 & \textbf{.7671} \\
    \arrayrulecolor{Note}
    \midrule
    \textbf{MODALS}$^\diamondsuit$ interpolation & \textbf{.7055} & .8137 & .7472 & .8104 & .7852 & .7725 & .7064 & .7630 \\
    \textbf{MODALS}$^\diamondsuit$ extrapolation & .6817 & .7913 & .7471 & .8136 & .7905 & .7824 & .7124 & .7599 \\
    \textbf{MODALS}$^\diamondsuit$ linear sampling & .6923 & .8114 & .7381 & .8090 & .7837 & .7745 & .6981 & .7582 \\
    \textbf{MODALS}$^\diamondsuit$ Gaussian noise & .6913 & \textbf{.8143} & .7422 & .8095 & .7788 & .7760 & .7180 & .7614 \\
    \textbf{EfficientCL}$^\blacklozenge$ PCA jittering & .6848 & .7893 & .7426 & .8141 & .7755 & .7854 & \textbf{.7200} & .7588\\
    \arrayrulecolor{black}
    \bottomrule
    \end{tabular}
    \caption{Performance on STS tasks for different augmentations with BERT as the backbone model. $^*$~is the reproduced result of unsupervised SimCSE~\cite{gao2021simcse}. The augmentations presented at the top part of the table are applied directly to sentences while the ones at the bottom part are added in the embedding space. $^\spadesuit$~denotes the results after adding the augmentation strategies proposed by~\citet{wei2019eda}. $^\heartsuit$~denotes the results after training with positive sentences generated via back-translation~\cite{sennrich2016improving}; we first translate the original text to French and then back to English. $^\clubsuit$~denotes the result produced by training with positive pairs generated by using a paraphrasing model based on PEGASUS~\cite{zhang2020pegasus}. $^\diamondsuit$~denotes the results after applying augmentation strategies proposed by~\citet{cheung2021modals}. $^\blacklozenge$~denotes the result produced by adding perturbation on the embedding layer with PCA jittering proposed by~\citet{ye2021efficient}.}
    \label{tab:sts_aug}
\end{table*}

We conducted experiments to assess the performance of various augmentation strategies (explained below) along with contrastive learning. The results are enumerated in~\tableautorefname~\ref{tab:sts_aug}. The strategies listed in the top part of the table are input-level augmentations, \ie they change the input sentence by applying transformations such as deleting or inserting words. The methods at the bottom are embedding-level augmentations. Paraphrasing provides the best results on average.

\paragraph{EDA.} Easy Data Augmentation (EDA)~\cite{wei2019eda} introduced 4 data augmentation strategies: synonym replacement using WordNet~\cite{miller1999wordnet}, random word insertion, random word position swapping, and random word deletion.

\paragraph{Back-translation.} Similar to paraphrasing, back-translation is a method that rephrases the sentence using pre-trained language models. It involves translating the sentence to another language and back. In our experiments, we use the pre-trained translation model between English and French by~\citet{tiedemann2020opus}.

\paragraph{MODALS.} \citet{cheung2021modals} proposed the following augmentation strategies at the embedding level: hard example interpolation (interpolate a sample with its closest hard example), hard example extrapolation (use the centre of a set of samples), linear sampling (perturb an embedding along the direction of two random samples), and randomly add Gaussian noise to the embeddings.

\paragraph{PCA jittering.}\citet{ye2021efficient} introduced this augmentation strategy that operates at the embedding level. It adds noise generated based on the application of PCA. 

\begin{algorithm*}[t]
    \begin{small}
    \caption{UNA for self-supervised pre-training}\label{alg:ssl_aug}
    \begin{algorithmic}[1]
    \algrenewcommand{\alglinenumber}[1]{\linenumbercolor\footnotesize#1}
    \State \textbf{Input}: A set of $N$ documents, $\mathcal{D}$
    \State Derive the TF-IDF representation of $\mathcal{D}$ and store it in matrix $\mathbf{Z}$
    \Comment \emph{as described in section~\ref{sec:UNA}}
    \State Store the $\max$ TF-IDF score (relevance) of each term in vector $\mathbf{s}$
    \For {every $\alpha$ batches}
    \Comment \emph{self-supervised pre-training}
    \For {document $j = 1$ to $B$}
    \Comment \emph{each batch has $B$ documents}
    \State The TF-IDF vector of document $j$ is given by $\mathbf{z} \subseteq$ row of $\mathbf{Z}$
    \Comment \emph{ignore terms not present in document $j$}
    \For{$i = 1$ to $n_z$}
    \Comment \emph{$n_z$ is the number of terms in document $j$}
    \State $p_i = \min\left(\beta\left(z_i-\min\left(\mathbf{z}\right)\right)/C, 1\right)$, where $C = \left(1/n_z\right)\sum_{i=1}^{n_z}\left(z_i-\min\left(\mathbf{z}\right)\right)$ for word replacement
    \State Determine the adjacent set of terms $\left(\mathcal{A}_i\right)$ to $i$ within radius $r$ based on the scores in $\mathbf{s}$
    \State With probability $p_i$ replace term $i$ with a term from $\mathcal{A}_i$; sample the replacement term w.r.t. the scores in $\mathbf{s}$
    \EndFor
    \State Generate document $j'$ from $j$ after the TF-IDF-driven stochastic term replacement
    \State Add document $j'$ to the set of UNA's hard negative samples for this batch, $\mathcal{H}_b$
    \EndFor
    \EndFor
    \State \textbf{Output}: $\mathcal{H}_b$ every $\alpha$ batches
    \end{algorithmic}
    \end{small}
\end{algorithm*}

\subsection{Results and analysis} 
\tableautorefname~\ref{tab:sts_aug} shows how established augmentation methods perform within a contrastive learning setup. We hypothesise that paraphrasing outperforms all other methods (on average) because it introduces more diverse but reliable positive pairs that assist in capturing useful representations. A drawback of employing pre-trained models for sentence rephrasing lies in the strong dependence on the specific pre-trained weights used. This seems to be significantly affecting some of the methods, such as back-translation. The embedding-level augmentations might facilitate a better generalisation (as they are not dependent on specific terms), but given that UNA operates on terms (as opposed to embeddings) using them in tandem with UNA did not outperform the paraphrasing and UNA combination (performance not shown).

\subsection{Implementation details of UNA} 
We implement UNA on our reproduced version of SimCSE~\cite{gao2021simcse} with PyTorch~\cite{paszke2019pytorch}. All experiments are conducted on 4 NVIDIA GeForce RTX 2080 Ti GPUs. UNA is applied during the self-supervised pre-training stage to create negative samples that contain similar words or sentence constructions to the anchor sentence. For clarity, we have also included UNA's algorithmic routine in Algorithm~\ref{alg:ssl_aug} (this is also described in section~\ref{sec:UNA}).

\begin{table}[t]
    \small
    \centering
    \begin{tabular}{ccc}
    \toprule
        $\alpha$ & \bf UNA & \bf Paraphrasing \& UNA \\
        \midrule
        9 &.8213            & .8289 \\
        7 &.8180            & .8295 \\
        5 &\textbf{.8235}   & \textbf{.8345} \\
        3 &.7973            & .8254 \\
        1 &.7929            & .8230 \\
        \arrayrulecolor{lightgray}
        \midrule
        \arrayrulecolor{black}
        0 &.8161            & .8235 \\
    \bottomrule
    \end{tabular}
    \caption{Performance on the STS-B development dataset (Spearman correlation) for different negative sample injection frequencies (every $\alpha$ training batches) during the training process. $\alpha=0$ denotes the absence of any form of augmentation.}
    \label{tab:nsteps}
\end{table}

We apply UNA once every $\alpha$ training batches. The downstream performance is influenced by the frequency ($\alpha$) of adding generated negative samples. If UNA is applied too frequently, the model may struggle to capture discriminative patterns between random negative samples. To find the optimal frequency, we conduct a grid search with $\alpha = \{1,3,5,7,9\}$ and evaluate performance on the STS-B development dataset (\tableautorefname~\ref{tab:nsteps}). The best performance is obtained by setting $\alpha=5$, and hence, this is what we use in our experiments.

\begin{table}[!t]
\small
    \centering
    \begin{tabular}{ccc}
    \toprule
        $\beta$ & \bf UNA & \bf Paraphrasing \& UNA\\
        \midrule
        0.8 & .8061             & .8328            \\
        0.7 & .8122             & .8354             \\
        0.6 & .8176             & \textbf{.8355}    \\
        0.5 & \textbf{.8235}    & .8345             \\
        0.4 & .7968             & .8320             \\
        0.3 & .7798             & .8313             \\
    \bottomrule
    \end{tabular}
    \caption{Performance on the STS-B development dataset (Spearman correlation) for different magnitudes of UNA controlled by hyperparameter $\beta$.}
    \label{tab:samplify}
\end{table} 

\begin{table}[!t]
  \small
  \centering
    \begin{tabular}{ccc}
        \toprule
        \bf $r$ terms & \bf UNA & \bf Paraphrasing \& UNA \\
        \midrule
        6{,}000 &.8101              & .8319  \\
        5{,}000 &.8158              & \textbf{.8363} \\
        4{,}000 &\underline{.8235}  & \underline{.8345}  \\
        3{,}000 &.8167              & .8332 \\
        2{,}000 &\textbf{.8264}     & .8322  \\
        1{,}000 &.8073              & .8331 \\
    \bottomrule
    \end{tabular}
    \caption{Performance on the STS-B development dataset (Spearman correlation) for different replacement term radius ($r$) settings. The best results are bolded and the second-to-best results are underlined.}
  \label{tab:size}
\end{table}

To determine the augmentation magnitude $\beta$, we validate values $\{0.3, 0.4, 0.5, 0.6, 0.7, 0.8\}$ on the STS-B development set (\tableautorefname~\ref{tab:samplify}). We set $\beta = 0.5$ as this setting yields the greatest average correlation score (across UNA and paraphrasing \& UNA).

Another hyperparameter in UNA is the radius $r$ for determining the set of terms in \textbf{Step 3}. Performance on the STS-B development set for different values of $r$ is shown in~\tableautorefname~\ref{tab:size}. The model reaches its peak performance for $r = 5{,}000$ and $2{,}000$ terms with and without paraphrasing, respectively. One potential explanation for this variation could be that the expanded paraphrasing vocabulary has an effect on the radius. We have decided to set $r = 4{,}000$ (with and without paraphrasing) which yields the second-best performance for both augmentation approaches.

\section{Complementary results}

\begin{table}[!ht]
\small
\centering
    \begin{minipage}{1.0\textwidth}
    \begin{tabular}{@{}rcc@{}}
        \toprule
        \textbf{Seed} & \textbf{SimCSE} & \textbf{Paraphrasing \& UNA} \\
        \midrule
        42      & .7532 & .7767 \\
        0       & .7526 & .7691 \\
        1       & .7452 & .7680 \\
        11      & .7457 & .7655 \\
        15      & .7488 & .7668 \\
        48      & .7478 & .7646 \\
        111     & .7651 & .7726 \\
        421     & .7395 & .7692 \\
        456     & .7562 & .7739 \\
        3407    & .7623 & .7707 \\
        \midrule
        \bf Mean (std.)    & .7516 (.0079) & .7698 (.0038) \\
        \bottomrule
    \end{tabular}
    \caption{Comparison on using different random seeds between SimCSE and SimCSE with the paraphrasing and UNA augmentations across the 7 STS tasks (average Spearman correlation) with BERT$_\texttt{base}$ model.}
    \label{tab:seed}
    \end{minipage}

    \vspace{20pt}
    
    \begin{minipage}{1.0\textwidth}
    \centering
    \setlength{\tabcolsep}{3pt}
    \begin{tabular}{@{}lcccccccc@{}}
    \toprule
    \multirow{2}{*}{\textbf{Augmentations}} & \multicolumn{8}{c}{\textbf{SentEval \textcolor{Line}{- STS tasks }}} \\
    \cmidrule(lr){2-9} 
    & \textbf{STS12} & \textbf{STS13} & \textbf{STS14} & \textbf{STS15} & \textbf{STS16} & \textbf{STS Bench.} & \textbf{SICK Rel.} & \textbf{Avg.} \\
    \midrule
    BERTbase (first-last avg.) & .3970 & .5938 & .4967 & .6603 & .6619 & .5387 & .6206 & .5670 \\
    BERTbase-whitening & .5783 & .6690 & .6090 & .7508 & .7131 & .6824 & .6373 & .6628 \\
    SCD-BERT$_\texttt{base}$~\cite{klein2022scd} & .6694 & .7803 & .6989 & .7873 & .7623 & .7630 & \underline{.7318} & .7419 \\
    Mirror-BERT$_\texttt{base}$~\cite{liu2021fast} & .691 & .811 & .730 & .819 & .757 & .780 & .691 & .754 \\
    DCLR-BERT$_\texttt{base}$~\cite{zhou2022debiased} & .7081 & .8373 & .7511 & .8256 & .7844 & .7831 & .7159 & .7722 \\
    AugCSE-BERT$_\texttt{base}$~\cite{tang2022augcse} & \underline{.7140} & \underline{.8393} & .7559 & \underline{.8359} & \underline{.7961} & .7961 & .7219 & \underline{.7798} \\
    DiffCSE-BERT$_\texttt{base}$~\cite{chuang2022diffcse} & \textbf{.7228} & \textbf{.8443} & \textbf{.7647} & \textbf{.8390} & \textbf{.8054} & \textbf{.8059} & .7123 & \textbf{.7849} \\
    Ours-BERT$_\texttt{base}$ & .6939 & .8158 & \underline{.7616} & .8304 & .7910 & \underline{.8030} & \textbf{.7412} & .7767 \\
    \midrule
    RoBERTabase (first-last avg.) & .4088 & .5874 & .4907 & .6563 & .6148 & .5855 & .6163 & .5657 \\
    RoBERTabase-whitening & .4699 & .6324 & .5723 & .7136 & .6899 & .6136 & .6291 & .6173 \\
    SCD-RoBERTa$_\texttt{base}$~\cite{klein2022scd} & .6353 & .7779 & .6979 & .8021 & .7729 & .7655 & \underline{.7210} & .7389 \\
    Mirror-RoBERTa$_\texttt{base}$~\cite{liu2021fast} & .666 & .827 & .740 & .824 & .797 & .796 & .697 & .764 \\
    DCLR-RoBERTa$_\texttt{base}$~\cite{zhou2022debiased} & .7001 & \underline{.8308} & \underline{.7509} & \textbf{.8366} & .8106 & .8186 & .7033 & .7787 \\
    AugCSE-RoBERTa$_\texttt{base}$~\cite{tang2022augcse} & .6930 & .8217 & .7349 & .8182 & \underline{.8140} & .8086 & .6877 & .7683 \\
    DiffCSE-RoBERTa$_\texttt{base}$~\cite{chuang2022diffcse} & \underline{.7005} & \textbf{.8343} & \textbf{.7549} & .8281 & \textbf{.8212} & \textbf{.8238} & .7119 & \textbf{.7821} \\
    Ours-RoBERTa$_\texttt{base}$ & \textbf{.7095} & .8188 & .7493 & \underline{.8334} & .8092 & \underline{.8189} & \textbf{.7352} & \underline{.7820} \\
    \bottomrule
    \end{tabular}
    \caption{Comparison with other pre-trained models on the SemEval STS tasks (Spearman correlation) with BERT and RoBERTa as the backbone model. Ours is presented with both UNA and paraphrasing added. The best results are bolded and the second-to-best results are underlined.}
    \label{tab:comp}
    \end{minipage}
\end{table}

\subsection{Random seeds}
To examine the robustness of UNA to random seeds, we conducted experiments by pre-training with $10$ random seeds on both SimCSE and UNA with paraphrasing. As shown in Table~\ref{tab:seed}, our approach is slightly more robust and yields consistently superior performance.

\subsection{Additional comparisons}
\label{app:comp}
We present a further comparison with additional pre-trained SSCL models as tabulated in~\tableautorefname~\ref{tab:comp}. Our method reached competitive results compared to SOTA methods with both backbone models, especially with RoBERTa.

\end{document}